\documentclass[10pt,twocolumn,letterpaper]{article}

\usepackage{iccv}
\usepackage{times}
\usepackage{epsfig}
\usepackage{graphicx}
\usepackage{amsmath}
\usepackage{amssymb}
\usepackage{textcomp}

\usepackage{color}
\usepackage{slashbox}
\usepackage{ruler}
\usepackage{booktabs}       
\usepackage{setspace}
\usepackage{algorithm}
\usepackage[noend]{algpseudocode}
\usepackage{dsfont}
\usepackage{array,multirow,graphicx}
\usepackage{float}

\usepackage[pagebackref=true,breaklinks=true,letterpaper=true,colorlinks,bookmarks=false]{hyperref}

\iccvfinalcopy 

\def\iccvPaperID{2511} 

\begin{document}

\title{Addressing Model Vulnerability to Distributional Shifts\\ over Image Transformation Sets}

\author{Riccardo Volpi\textsuperscript{1}, Vittorio Murino\textsuperscript{1,2,}\thanks{VM is also with Huawei Technologies (Ireland) Co., Ltd., Dublin.}\\
{\normalsize \textsuperscript{1}Istituto Italiano di Tecnologia \textsuperscript{2}Universit\`a di Verona}\\
{\small \texttt{\{riccardo.volpi,vittorio.murino\}@iit.it}}
}

\maketitle
\begin{abstract}
We are concerned with the vulnerability of computer vision models to distributional shifts. We formulate a combinatorial optimization problem that allows evaluating the regions in the image space where a given model is more vulnerable, in terms of image transformations applied to the input, and face it with standard search algorithms. We further embed this idea in a training procedure, where we define new data augmentation rules according to the image transformations that the current model is most vulnerable to, over iterations. An empirical evaluation on classification and semantic segmentation problems suggests that the devised algorithm allows to train models that are more robust against content-preserving image manipulations and, in general, against distributional shifts\footnote{Code at~{\scriptsize \url{github.com/ricvolpi/domain-shift-robustness}}}.
\end{abstract}

\section{Introduction}\label{sec:intro}
When designing a machine learning system, we generally desire it may perform well on a wide realm of different domains. However, the training data at disposal is typically defined by samples from a limited number of distributions, resulting in unsatisfactory performance when the model has to process data from unseen distributions~\cite{Daume2006,Blitzer2006,BenDavid2006,NameTheDataset}. This problem is typically referred to as \textit{distributional shift} or \textit{domain shift}, and it was shown to affect models even in cases where the distance between training and testing domain is---apparently---very limited~\cite{recht2018cifar,recht2019imagenet}.

This vulnerability also affects the robustness of machine learning models against input manipulations~\cite{gilmer2018motivating,hosseini2017negative,hosseini2018cvprw}, potentially leading to harmful situations. As a concrete example, consider the algorithms that analyze images uploaded to social networks in order to evaluate, \eg, if an image contains violence or adult content. The huge set of image modifications that users might carry out can make the underlying learning systems fail in several ways if they, accidentally or with malicious intent, cause a \textit{shift} that the models are not able to figure out. 
Recognizing this weakness of modern learning systems, an important research direction is defining methods to understand \textit{a priori} which distributional shifts will lead to a fail of the model. 

In this paper, we start from this idea, and develop methods to evaluate and improve the performance of machine learning models for vision tasks, when the input can be modified through a series of content-preserving image transformations. By ``content-preserving''~\cite{gilmer2018motivating}, we intend transformations that do not modify an image content, but only the way it is portrayed (\eg, modifying RGB intensities, enhancing contrast, applying filters, etc). 

\begin{figure}[!t]
\centering
\includegraphics[width=0.725\linewidth]{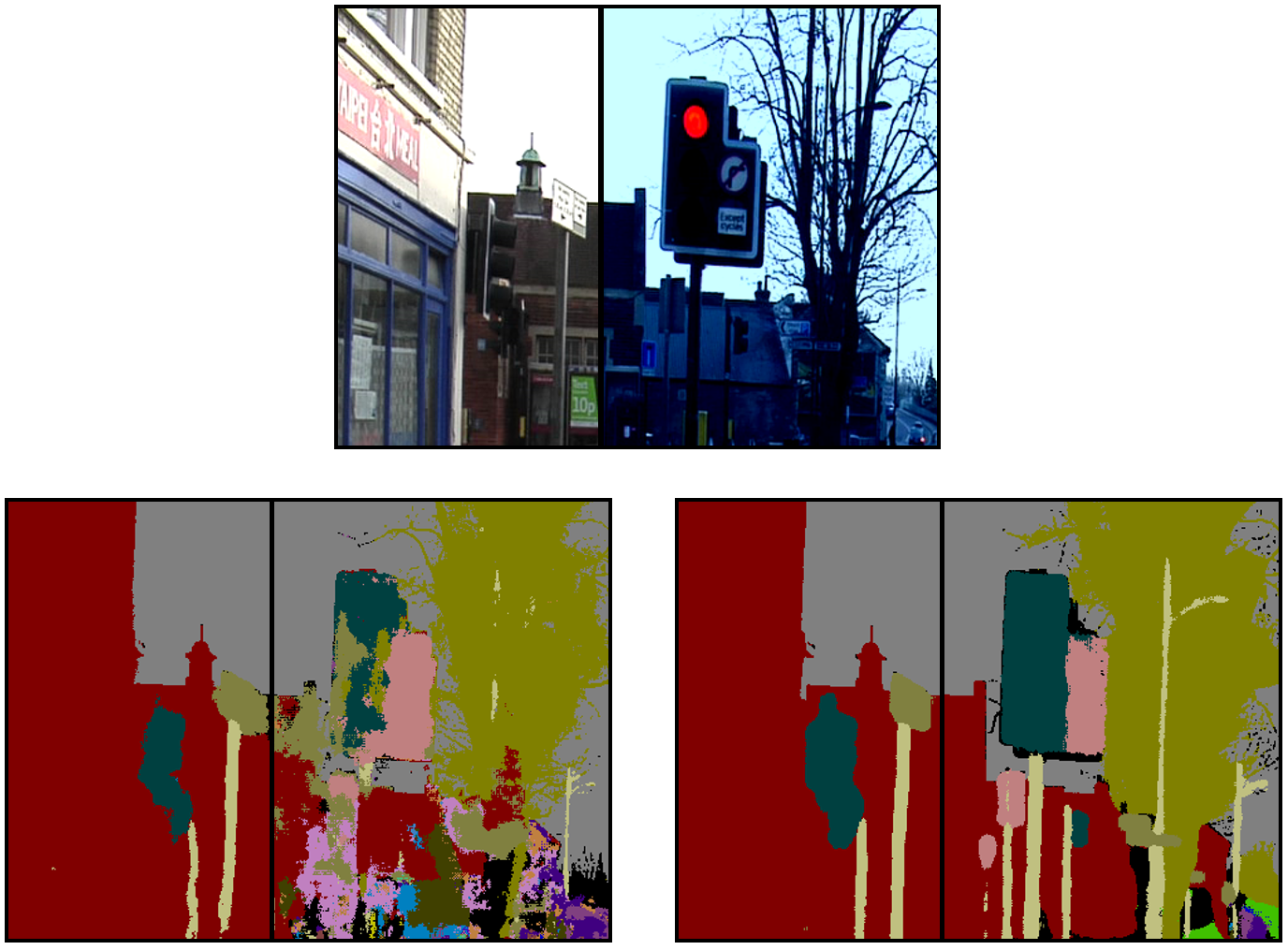}
\caption{\footnotesize \textit{(Top)} Image subdivided in 2 parts, in which the left part is original, and the right part was subject to a content-preserving (appearance) transformation. Image transformations 
can cause distributional shifts that models are not able to handle \textit{(bottom-left)}. Models trained with the methods proposed in this paper are more robust against a variety of image transformations \textit{(bottom-right)}}
\label{fig:FIG1}
\end{figure}

We cast this problem in terms of combinatorial optimization. Given a black-box model, a bunch of samples, and a set of image transformations, our goal is to individuate the distributional shifts that the model is most vulnerable to when image transformation tuples (namely, concatenations of transformations) are applied. To find these tuples, we investigate two different search algorithms---random search and evolution-based search---showing that it is easy to find tuples that severely deteriorate the model performance for a variety of tasks, such as face detection, semantic scene segmentation, and classification. 
The main application for this method as-is, is to evaluate the vulnerability of a machine learning model \textit{before} its deployment. Coupled with proper transformation sets, this tool can indeed be used to verify the robustness of a model under a broad variety of input manipulations and visual conditions.\\

Furthermore, we introduce a training procedure to learn more robust models against this class of transformations. We design an algorithm where new data augmentation rules are included over iterations, in order to cover the distributional shifts where the current model is more vulnerable. We show that models trained in this way are more robust against content-preserving input image manipulations and, moreover, better generalize to unseen scenarios at test time. 



\subsection{Background and related work}\label{rel_work}

\paragraph{Vulnerability of learning systems.}

Recently, the vulnerability of learning systems in different scenarios has gained a lot of attention, in particular in relations with computer vision models (typically, deep convolutional neural networks, or ``ConvNets''~\cite{LeNet}). A widely studied area is the one related to defense against adversarial perturbations. Gilmer et al.~\cite{gilmer2018motivating} makes a distinction between adversarial samples that are merely \textit{content-preserving} or also  \textit{indistinguishable} from the originals. The latter case takes into account imperceptible (to human eye) input perturbations that make a model fail. This paradigm has been extensively studied in a substantial body of works (\eg, ~\cite{FastGradientMethod,Certifiable,LeeRa17,Heinze-DemlMe17,moosavi2017universal,madry2018pgd}). 

Instead, we can include a broader range of transformations in the ``content-preserving'' class. Given some input, the content-preserving transformations are the ones that do not change its content, even if the appearance may change significantly. For example, Gilmer et al.~\cite{gilmer2018motivating} explore the performance of a classifier trained on MNIST~\cite{MNIST} when the input is modified by altering the background or adding random lines. Brown et al.~\cite{brown2017patch} show that we can cause model failure by including adversarial patches in an image. Hosseini et al.~\cite{hosseini2017negative} realize that vision models are vulnerable to negative images. The same research group, in an other work~\cite{hosseini2018cvprw}, shows that we can find hue and saturation shifts for a given image that a model is vulnerable to. Furthermore, recent works~\cite{hendrycks2018benchmarking,geirhos2018imagenettrained} show that state-of-the-art ImageNet~\cite{ImageNet} models are vulnerable towards simple image modifications. In particular, Hendrycks et al.~\cite{hendrycks2018benchmarking} have found that these models are not resistant towards basic noise sources, and Geirhos et al.~\cite{geirhos2018imagenettrained} have shown that these models are biased towards the texture of the objects.

As stated by Gilmer et al.~\cite{gilmer2018motivating}, we also deem that ``\textit{the space of content-preserving image transformations remains largely unexplored in the literature}''.
One of the aims of this work is to help filling this gap, proposing methods to study, generate, and be robust against content-preserving image transformations. Differently from previous works~\cite{gilmer2018motivating,hosseini2018cvprw}, we are not interested in finding adversarial transformations for \textit{single} images. We are instead interested in discovering the distributional shifts that a model is \textit{in general} more vulnerable, applying the same transformation to all the images in the provided set. In this sense, this work is related to Moosavi-Dezfooli et al.~\cite{moosavi2017universal}, where a single, imperceptible perturbation that fools ImageNet models is found. 

\paragraph{Robustness against distributional shifts.}

There is a significant body of works whose goal is overcoming issues related to distributional shift. 

One of the main research direction is \textit{domain adaptation}~\cite{Daume2006,Blitzer2006,Saenko2010,Ganin,ADDA,DeepCORAL,morerio2018,volpi2018cvpr}, where the goal is to better generalize on domains of interest for which only unlabeled data are available. While there are algorithms that tackle this problem with remarkable results across a variety of tasks, the assumption of an \textit{a priori} fixed target distribution is often too strong. In \textit{domain generalization}~\cite{li2017iccv,motiian2017iccv,muandet2013icml,DG3,Mancini2018,volpi2018nips,li2019arxiv1,li2019arxiv2,volpi2018nips} the problem of dealing with unseen distributions is coped. Usually, the proposed algorithms start from the assumption that the training dataset comprises a number of different populations. One exception is the method proposed by Volpi et al.~\cite{volpi2018nips}, where the authors introduce a worst-case formulation that improves generalization performance across distributions close to the training one in the semantic space, using a single-source distribution as starting point. Tobin et al. \cite{DomainRandomization} introduce \textit{domain randomization} for models trained through simulated data. It generates a randomized variety of visual conditions during training, hoping to better generalize when coping with real data. 

In this context, the method devised by Volpi et al.~\cite{volpi2018nips} is the most related approach to the proposed training strategy (detailed in Section~\ref{sec:robust_train}) since they are aimed at learning models that better generalize to unseen scenarios, without any assumptions on the number of data populations in the training set. As results will show, the competing algorithm~\cite{volpi2018nips} results in models that are only slightly more resistant than the Empirical Risk Minimization (ERM) baseline in the testbed presented in Section~\ref{sec:search_algos}, and significantly less performing than models trained through the Algorithm proposed in this work in domain generalization settings. 

\section{Problem formulation}\label{sec:prob_form}

Let $\mathbb{M}$ be a model that takes in input images and provides an output according to the given task. Let $D = \{(x^{(i)}, y^{(i)})\}_{i=0}^{N_D} \sim P(X,Y)$ be a set of datapoints with their labeling, drawn from some data distribution. Finally, let $\mathbb{T} = \{(\tau^{(j)}, \l_j^{(k)}), j=0\dots N_T, k=0\dots N_j\}$ be a set where each object $t=(\tau^{(j)}, \l_j^{(k)})$ is a data transformation $\tau$ with a related magnitude $\l$. The transformations give in output datapoints in the same format as the input ones (RGB images throughout this work)\footnote{To provide a practical example, one object from $\mathbb{T}$ might be the \textit{``brightness''} operation, and the intensity level might be $+6\%$.}. The transformations can be concatenated and repetitions are allowed; we define a composite transformations as a transformation tuple. We define the set of all the possible transformation tuples that one can obtain by combining objects in $\mathbb{T}$ as

\vspace{-3pt}
\begin{equation}\label{eq:tuples}
\small
    \mathbb{T}_{N} = \{ \text{all $N-$tuples from $\mathbb{T}$} \}
\end{equation}

A tuple $T \in \mathbb{T}_N$ is the concatenation of $N$ objects from $\mathbb{T}$, and we define it as $T = (t_{1}, \dots, t_{N})$, with $t_n \in \mathbb{T}$. When we apply the tuple $T$ to a datapoint $x$, we apply all the transformations from $t_{1}$ to $t_{N}$. Armed with this set, we propose the following combinatorial optimization problem
 
\vspace{-5pt}
\begin{equation}\label{eq:worstcase}
\small
    \min_{T^* \in \mathbb{T}_{N}} f\big(\mathbb{M}, T^*, D\big) 
\end{equation}

where $f$ is a fitness function that measures the performance of a model $\mathbb{M}$ when provided with some labelled datapoints $D$, transformed according to the tuple $T^*$. Assuming that the maximum and minimum values for the metric associated with $f$ are $1$ and $0$, respectively, we have 

\vspace{-3pt}
\begin{equation}\label{eq:mapping}
\small
    f: T \longrightarrow [0,1] \subset \mathbb{R}^1 \nonumber 
\end{equation}

Intuitively, the $N-$tuples that induce lower $f$ values, are the ones that a model $\mathbb{M}$ is more vulnerable to, with respect to the chosen metric. For classifiers, the optimization problem~\ref{eq:worstcase} assumes the form

\vspace{-8pt}
\begin{equation}\label{eq:worstcase_class}
\small
    \min_{T^* \in \mathbb{T}_{N}} f := \frac{1}{N_D} \sum^{N_D}_{i=0} \mathds{1}\big\{y^{(i)} = \mathbb{M}\big(T^*(x^{(i)})\big)\big\} 
\end{equation}

In general, one can define an instance of problem~\ref{eq:worstcase} if provided with a set of annotated samples $D$, a transformation set $\mathbb{T}$ (and, consequently, a tuple set \textbf{$\mathbb{T}_N$}), a model $\mathbb{M}$, a measure to evaluate the performance of the model, and, consequently, a fitness function $f$. It is not required to have access to the model parameters: it can be a black-box. A legit critique to this formulation is that we are not constraining the transformation tuples to be content-preserving. For instance, in the classification problem~\ref{eq:worstcase_class}, a proper formulation would include a constraint similar to the following:   

\vspace{-6pt}
\begin{equation}\label{eq:worstcase_oracle}
\small
    \mathcal{O}\big(x^{(i)}\big) = \mathcal{O}\big(T(x^{(i)})\big) \text{\phantom{A}} \forall{i}, \nonumber
\end{equation}

which means that an oracle $\mathcal{O}\big(.\big)$ would classify the transformed images in the same way as the original images. In this work, we do not explicitly constraint the transformation tuples to be content-preserving through the optimization problem. We satisfy the constraint by properly defining the set of available image transformations, \eg, focusing on simple color transformations such as RGB enhancement, contrast/brightness adjustments, and setting a proper value for $N$ in $\mathbb{T}_N$. Explicitly imposing the constraint is an important research direction, since it would allow to consider more complex sets, and we reserve it for future work.

\subsection{Transformation set and size of the search space}\label{sec:search_space}

Given a transformation set $\mathbb{T}$ with $N_T$ available transformations $\tau^{(j)}$, where the $j_{th}$ has $N_j$ available magnitude values, the size of $\mathbb{T}_N$, and consequently the size of the search space of the optimization problem~\ref{eq:worstcase}, is $S=(\sum_{j=0}^{N_T}{N_j})^N$. 

In this work, we consider a transformation set $\mathbb{T}$ including standard image transformations from the Python library Pillow~\cite{pil-library}, as done by Cubuk et al.~\cite{cubuk2018autoaugment}, and a few more we included. It is defined by the following transformations, with the number of available intensity levels indicated in parenthesis: \textit{autocontrast (20), sharpness (20), brightness (20), color (20), contrast (20), grayscale conversion (1), R-channel enhancer (30), G-channel enhancer (30), B-channel enhancer (30), solarize (20)}. The description of the various transformations is reported in Appendix~\ref{appendixA}, as well as the ranges of intensity levels. This set results in a search space with size $S=211^N$. Throughout this work, we will consider tuple sets $\mathbb{T}_N$ with $N=3$ and $N=5$, resulting in search spaces with size in the order of $\sim10^6$ and $\sim10^{12}$, respectively.  

\section{Searching worst-case image transformations}\label{sec:search_algos}

In this section, we analyze different solutions to face the combinatorial optimization problem~\ref{eq:worstcase}. Specifically, the two approaches rely on random search and evolution-based~\cite{mitchell1998iga} search. We provide a proof of concept experiment on MNIST models, and report a more exhaustive experimental evaluation in Section~\ref{sec:experiments}.  

\subsection{Random search.} 

Facing the optimization problem~\ref{eq:worstcase} through random search is important for several reasons. First, it is the simplest approach that we can adopt, thus it is worth to be explored. Further, random search is often a very strong baseline to compare against, as shown, \eg, in hyper-parameter optimization~\cite{bergstra2012jmlr} and neural architecture search~\cite{li2019nas}. Finally, it sheds light on a relevent question: \textit{how is a model affected by random image transformations?}

The idea is to evaluate the fitness function $f$ over an arbitrary number of random transformation tuples, thus the implementation is straightforward. For clarity and reproducibility, we detail it step-by-step on Algorithm~\ref{alg:random_search}. In the following, we will refer to this procedure as RS (short for Random Search)

\begin{algorithm}[t]
\caption{RS (Random Search)}
\label{alg:random_search}
\begin{spacing}{1.1}
\begin{algorithmic}[1]
\small
\State \textbf{Input:} N-tuple set $\mathbb{T}_N$, model $\mathbb{M}$, dataset $D = \{x^{(i)},y^{(i)}\}_{i=1}^{N_D}$, fitness function $f$.
\State \textbf{Output:} transformation $T \in \mathbb{T}_N$
\State \textbf{Initialize:} $f_{min} \gets 1.$
\For{$k=1,...,K$}
\State Sample $T^*$ uniformly from $\mathbb{T}_N$
\State $f^* \gets f(\mathbb{M}, T^*, D)$
\If{$f^* < f_{min}$}
\State $T \gets T^*$
\EndIf
\EndFor
\end{algorithmic}
\end{spacing}
\end{algorithm}

\subsection{Evolution-based search.} 

We define a simple genetic algorithm~\cite{mitchell1998iga}, aimed at minimizing the objective in problem~\ref{eq:worstcase}. Each individual of the population is defined by a transformation $N-$tuple from a set $\mathbb{T}_N$. We define standard \textit{Selection}, \textit{Crossover} and \textit{Mutation} operations. For a detailed explanation of genetic algorithms and the definitions we provided, we refer to~\cite{mitchell1998iga}. In the following, we briefly discuss how we use these concepts in our framework.  

\begin{itemize}
    \item \textbf{Selection.} Given in input a population \texttt{pop}$=\{ T^p \}_{p=1}^P$, the fitness score of each individual \texttt{fit}$=\{ f^p \}_{p=1}^P$, and a positive integer $\hat{P}$, returns in output a population of $\hat{P}$ individuals sampled from \texttt{pop} with individual probabilities proportional to $\frac{1}{f^p}$.
    
    \item \textbf{Crossover.} Given in input two initialized populations \texttt{pop1}$=\{ T^p \}_{p=1}^P$, where $T^p = (t_1^p, \dots, t_N^p)$ and \texttt{pop2}$=\{ \tilde{T}^p \}_{p=1}^P$, where $\tilde{T}^p = (\tilde{t}_1^p, \dots, \tilde{t}_N^p)$, for each couple of elements $\{(T^p,\tilde{T}^p)\}_{p=1}^P$ we uniformly draw an integer $n \in [1,N]$ and return the following two individuals: $T^{p,1} = (t_1^p, \dots, t_n^p, \tilde{t}_{n+1}^p, \dots, \tilde{t}_N^p)$ and $T^{p,2} = (\tilde{t}_1^p, \dots, \tilde{t}_n^p, t_{n+1}^p, \dots, t_{N}^p)$. The output is the population defined by the $2P$ new individuals. 
    
    \item \textbf{Mutation.} Given in input an initialized population \texttt{pop1}$=\{ T^p \}_{p=1}^P$ and a mutation rate $\eta$, it changes each transformation of each tuple in \texttt{pop} with probability $\eta$, sampling from $\mathbb{T}$.
    
\end{itemize}

Endowed of these methods, we implement an evolution-based search procedure, detailed in Algorithm~\ref{alg:evolution_search}. The complexity is $\mathcal{O}(PK)$, where $P$ is the population size and $K$ is the number of evolutionary steps. Notice that the operations associated with lines 5 and 11, namely computing the fitness function value for each transformation in the population, constitute the computationally expensive part of the algorithm. For each run, we perform $P(K + 1)$ fitness function evaluations. In the following, we will refer to this procedure as ES (short for Evolution-based Search).

{
\begin{algorithm}[t]
\caption{ES (Evolution-based Search)}
\label{alg:evolution_search}
\begin{spacing}{1.1}

\begin{algorithmic}[1]

\small

\State \textbf{Input:} N-tuple set $\mathbb{T}_N$, model $\mathbb{M}$, dataset $D = \{x^{(i)},y^{(i)}\}_{i=1}^{N_D}$, fitness function $f$, population size $P$, mutation rate $\eta$.
\State \textbf{Output:} transformation $T \in \mathbb{T}_N$



\State \textbf{Initialize:} $f_{min} \gets 1.$
\State {\color{white}\textbf{Initialize:}} \texttt{pop} $\gets$ $\{$T$^p\}_{p=1}^P$ sampling from $\mathbb{T}_N$
\State {\color{white}\textbf{Initialize:}}  \texttt{fit} $\gets \{f^{p}  \gets f(\mathbb{M}, T^p, D)\}_{p=1}^P $

\For{$k=1,...,K$}

\State \texttt{newpop1}  $\gets$  \textbf{Select}(\texttt{pop}, \texttt{fit}, $\frac{P}{2}$)
\State \texttt{newpop2} $\gets$ \textbf{Select}(\texttt{pop}, \texttt{fit}, $\frac{P}{2}$)
\State \texttt{pop} $\gets$ \textbf{Crossover}(\texttt{newpop1}, \texttt{newpop2})
\State \texttt{pop} $\gets$ \textbf{Mutation}(\texttt{pop},$\eta$)
\State \texttt{fit} $\gets \{f^{p}  \gets f(\mathbb{M}, T^p, D)\}_{p=1}^P $
\For{$p=1,...,P$} 
\If{\texttt{fit}$[p] < f_{min}$}
\State $T \gets \texttt{pop}[p]$
\EndIf
\EndFor
\EndFor
\end{algorithmic}

\end{spacing}
\end{algorithm}
}

\subsection{Proof of concept: MNIST}\label{sec:test_mnist}

The MNIST dataset~\cite{MNIST} is defined by $28\times 28$ pixel images, representing white digits on a black background. It is divided into a $50,000$ sample training set and a $10,000$ sample test set. In our experiments, we train a small ConvNet (\textit{conv-pool-conv-pool-fc-fc-softmax}) on the whole training set, via backpropagation~\cite{BackProp}. We resize the images to $32\times 32$ pixels, in order to be comparable with other digit datasets (in view of the domain generalization experiments reported in Section~\ref{sec:experiments}). We apply the search algorithms (RS and ES) on problem~\ref{eq:worstcase} using $1,000$ samples from the test set. We set $N=3$, namely, we use transformation tuples defined by three transformations.

\begin{figure}[!t]
\centering
\includegraphics[width=0.9\linewidth]{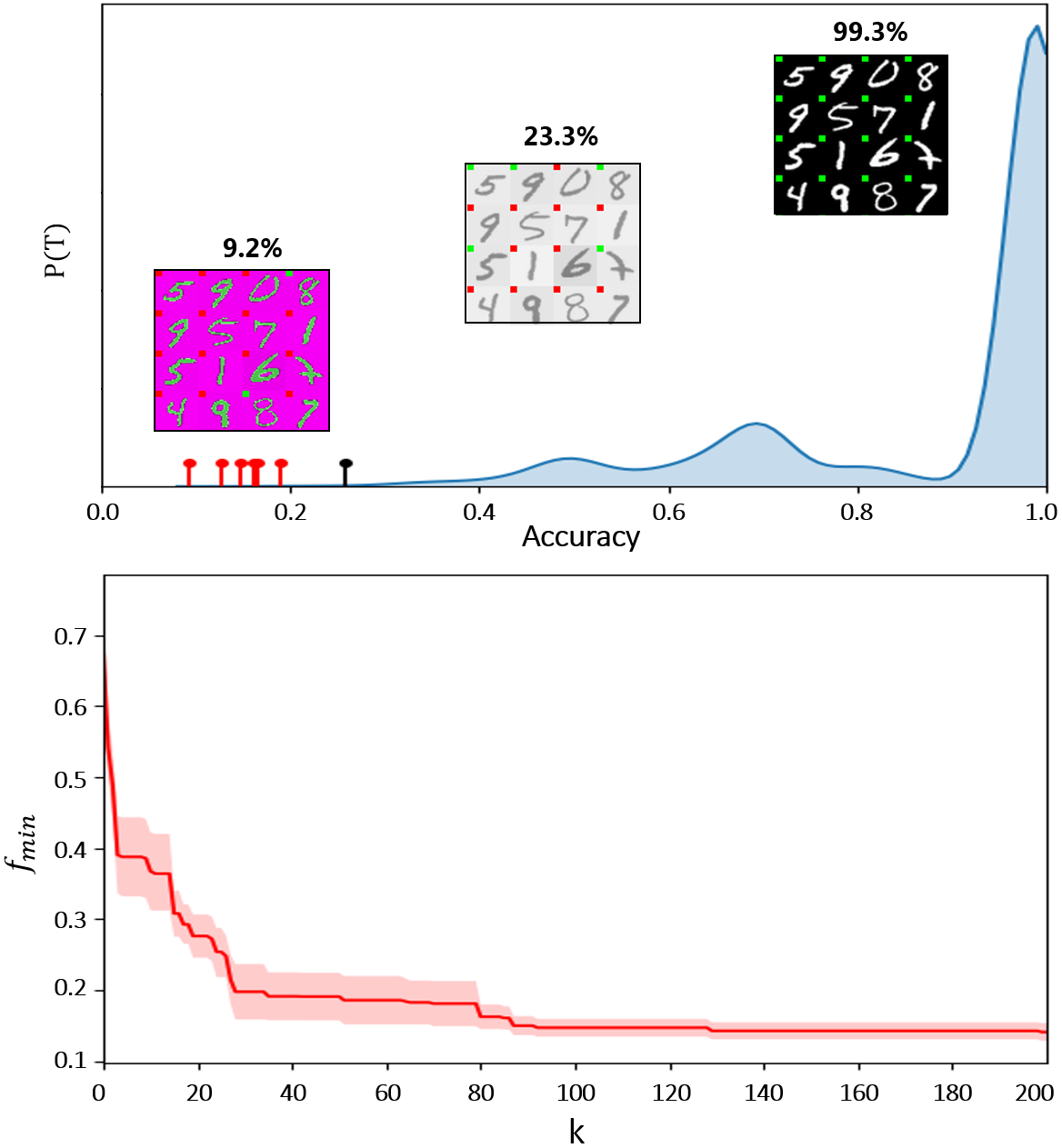}
\caption{\footnotesize \textit{Top.} The blue curve represents the density plot related to $10,000$ transformation tuples uniformly sampled for RS. The black flag represents the $0.1\%$ probability threshold for RS. The red flags represent the accuracy values obtained via ES. Three examples of sample sets resulting from different transformations are reported, with green and red squares indicating whether the sample is mis-classified or not, respectively. \textit{Bottom.} Evolution of $f_{min}$ value averaged over $6$ runs of ES.}
\label{fig:mnist_transf}
\end{figure}

The blue curve in Figure~\ref{fig:mnist_transf} is the density plot associated with all the fitness function values obtained while running RS for $K=10,000$ iterations, using as model $\mathbb{M}$ the trained ConvNet---that achieves $99.3\%$ accuracy on the clean test set. The accuracy values are reported on the x-axis. Values lower than the one indicated by the black flag have less than $0.1\%$ probability to be achieved by transforming the input through transformation tuples sampled from $\mathbb{T}_N$. This plot provides a glance on the vulnerability of MNIST models to the image transformations included in our set. It shows that there is a substantial mass of transformation tuples that the model is resistant to, but, even though with lower probability to be sampled, there are transformation tuples against which the model is severely vulnerable. 

Table~\ref{tab:worst_case_mnist} (RS row) shows the minimum accuracy obtained in $10,000$ evaluations of the fitness function $f$, averaged over $6$ different models. We report results associated with both models trained via standard ERM (homonymous column) and models trained through the method proposed by Volpi et al.~\cite{volpi2018nips} (\textit{``ADA''} column). 
As one can observe, both types of models are severely vulnerable to the transformation tuples found through RS. For comparison with previous work, we also report results obtained on negative images~\cite{hosseini2017negative} and on images with random hue/value perturbations~\cite{hosseini2018cvprw} (for the latter, we used the original code). 

\begin{table}[!t]
\begin{center}
{\scriptsize
\setlength{\tabcolsep}{2.5pt}
\begin{tabular}{@{}rccccc@{}}
 \multicolumn{6}{c}{\footnotesize \textbf{Performance of MNIST models}} \\
\toprule
& \multicolumn{5}{c}{\textbf{Training procedure}} \\
\cmidrule(r){2-6}
\textbf{Test} & \textbf{ERM} & \textbf{ADA}~\cite{volpi2018nips} & 
\textbf{RDA} & \textbf{RSDA} & \textbf{ESDA}  \\
\midrule

Original & $.993 \pm .001$ & $.992 \pm .001$ & $.993 \pm .001$ & $.992 \pm .001$ & $.993 \pm .001$ \\
\midrule

RS 
& $.160 \pm .024$ & $.192 \pm .025$ & $.941 \pm .011$ & $.977 \pm .001$ & $.979 \pm .003$ \\
\midrule

ES 
& $.122 \pm .028$ & $.177 \pm .042$ & $.927 \pm .007$ & $.979 \pm .005$ & $.977 \pm .000$ \\

\midrule

Neg.~\cite{hosseini2017negative} & $.436 \pm .042$ & $.448 \pm .046$ & $.991 \pm .001$ & $.992 \pm .001$ & $.992 \pm .001$ \\

\midrule

SAE~\cite{hosseini2018cvprw} & $.979 \pm .005$ & $.980 \pm .005$ & $.985 \pm .004$ & $.974 \pm .007$ & $.971 \pm .016$ \\

\bottomrule

\end{tabular}
} 
\end{center}
\caption{\footnotesize Accuracy values associated with MNIST models in different training/testing conditions, averaged over $6$ different training runs. Each row is associated with a different test: \textit{Original} refers to performance achieved on clean test samples; \textit{RS} and \textit{ES} refer to results obtained applying the transformations found via RS and ES, respectively; \textit{Neg} and \textit{SAE} are related to adversarial attacks detailed in~\cite{hosseini2017negative} and~\cite{hosseini2018cvprw}, respectively. Each column is related to models trained with a different procedure.} 
\label{tab:worst_case_mnist}
\end{table}







We proceed by approaching problem~\ref{eq:worstcase} through ES, setting population size $P=10$, number of generations $K=99$ and mutation rate $\eta=0.1$. With this setting, the number of fitness evaluations is $1,000$. The red flags in Figure~\ref{fig:mnist_transf} indicate the $f_{min}$ values achieved on $6$ different runs, using the same ConvNet as in the RS experiment. A comparison between the $0.1\%$ threshold (black flag) and the results obtained via evolution shows that ES allows to efficiently find low-probability transformation tuples that the model is most vulnerable to. Furthermore, even though we set $K=99$, EF can find transformation tuples that go beyond the $0.1\%$ threshold in less iterations. We report this evidence in Figure~\ref{fig:mnist_transf} (bottom), which shows the evolution of $f_{min}$ during $200$ iterations of ES. Comparing this result with the ones pictured in Figure~\ref{fig:mnist_transf} (top), one can observe that even by setting $K\simeq30$ ($310$ fitness function evaluations), ES outperforms the $0.1\%$ threshold in the RS results. We report numerical results in Table~\ref{tab:worst_case_mnist} (ES row), where we average over $6$ models the lowest $f_{min}$ achieved over $6$ runs of ES with different initializations. In Section~\ref{sec:experiments}, we will provide a more exhaustive analysis of the efficacy of RS and ES to approach different instances of problem~\ref{eq:worstcase}. 

\section{Training more robust models}\label{sec:robust_train}

{
\begin{algorithm}[t]
\caption{Robust Training}
\label{alg:robust_train}
\begin{spacing}{1.1}
\begin{algorithmic}[1]
\small
\State \textbf{Input:} $D=\{x^{(i)},y^{(i)}\}_{i=1}^{N_D}$, initialized weights $\theta_0$, $N-$tuple set $\mathbb{T}_N$, initialized data augmentation set $\mathbb{T}_{tr}$, learning rate $\alpha$.
\State \textbf{Output:} learned weights $\theta$
\State \textbf{Initialize:} $\theta \gets \theta_0$
\For{$h=1,...,H$}
\For{$j=1,...,J$}
\State Sample $(x, y)$ uniformly from $D$
\State Sample $T$ uniformly from $\mathbb{T}_{tr}$
\State $\theta \gets \theta - \alpha \nabla_{\theta} \ell(\theta; (T\{x\}, y))$
\EndFor
\State Find $T^* \in \mathbb{T}_N$ by running RS or ES on a subset of $D$
\State Append $T^*$ to $\mathbb{T}_{tr}$
\EndFor
\While{training is not done}
\State Sample $(x, y)$ uniformly from dataset
\State Sample $T$ uniformly from $\mathbb{T}_{tr}$
\State $\theta \gets \theta - \alpha \nabla_{\theta} \ell(\theta; (T\{x\}, y))$
\EndWhile
\end{algorithmic}
\end{spacing}
\end{algorithm}
}

In this section, we detail two straightforward methods devised to train models that are robust against content-preserving transformations from a given set. 

The simplest approach that one can devise is likely the following: given a set $\mathbb{T}_N$, we can perform data augmentation by sampling transformation tuples $T \in \mathbb{T}_N$ and applying them to the training images throughout the training procedure. We term this method \textit{Randomized Data Augmentation}, in short RDA. This technique can be interpreted as an application of domain randomization~\cite{DomainRandomization} to real data instead of simulated ones.

Drawing inspiration from the literature related to adversarial robustness~\cite{FastGradientMethod,Certifiable,volpi2018nips}, where a loss is minimized with respect to adversarially perturbed inputs, we devise a method that is more effective than RDA in our setting. We propose a training procedure where transformation tuples that the current model is most vulnerable to are searched throughout the training procedure (via RS or ES), and data augmentation is performed according to the so-found transformations. We implement this idea as follows: (a) we define a transformation set to sample from during training (the ``data augmentation set'' $\mathbb{T}_{tr}$), that at the beginning of training only comprises the \textit{identity} transformation; (b) we train the network via gradient descent updates~\cite{cauchy1947gradient}, augmenting samples by applying transformations uniformly sampled from $\mathbb{T}_{tr}$ (in this work, the loss $\ell$ used is the cross-entropy function between the output of the model and the ground truth labels); (c) we run RS or ES, using appropriate fitness function $f$ and tuple set $\mathbb{T}_N$, and append the so-found transformation tuple to $\mathbb{T}_{tr}$. We alternate between steps (b) and (c) for the desired number of times, and (d) we repeat step (b) until the value of the loss $\ell$ is satisfactory. The procedure is also detailed in Algorithm~\ref{alg:robust_train}. 

As results will show, the latter method performs significantly better than RDA in several settings. In the next Sections, we will refer to this method as RSDA or ESDA, short for \textit{Random Search Data Augmentation} and \textit{Evolution-based Search Data Augmentation}, respectively

\section{Experiments}\label{sec:experiments}

In Section~\ref{sec:test_mnist}, we provided a first evidence that the problem formulation introduced in Section~\ref{sec:prob_form} can be useful to detect harmful distributional shifts for a given model---in terms of image transformations. In Section~\ref{sec:robust_train}, we introduced different methods to train more robust models. 

In this section, we further validate the effectiveness of RS and ES on different instances of problem~\ref{eq:worstcase}, associated with models for classification, semantic segmentation and face detection. Furthermore, we evaluate the performance of classification and semantic segmentation models trained through RDA, RSDA and ESDA, assessing both their robustness against image transformations and their domain generalization properties. When we search for transformations while running RSDA and ESDA (Algorithm~\ref{alg:robust_train}, line 9), we set $K=100$ for RS and $K=10$ for ES. When we apply ES, we set number of individual $P=10$ and mutation rate $\eta=0.1$ throughout the entire analysis. We use accuracy as evaluation metric in all the experiments.  

\subsection{Digit Recognition}

\paragraph{Experimental setup.}

We adopt the same experimental setting detailed in Section~\ref{sec:test_mnist}. We train models via ERM and RDA for $10^6$ gradient descent updates. When we train models through RSDA/ESDA, we set $J=10^4$ and $H=100$, running a total of $10^6$ weight updates also in this case. We use a subset of $1,000$ samples from the training set when we run RS/ES (Algorithm~\ref{alg:robust_train}, line 9). In all the experiments, we set the size of the transformation tuples as $N=3$. We use Adam~\cite{AdamOptimizer} as optimizer, with learning rate $\alpha=3 \cdot 10^{-4}$, $\beta_1=0.9$, $\beta_2=0.999$ and $\epsilon=10^{-8}$. 

In addition to assessing model vulnerability against the transformations found via RS and ES, we also evaluate the domain generalization capabilities of MNIST models, testing on different, unseen digit datasets (SVHN~\cite{SVHN}, SYN~\cite{Ganin}, MNIST-M~\cite{Ganin}, USPS~\cite{USPS}), following the evaluation protocol used by Volpi et al.~\cite{volpi2018nips}. Samples from every dataset were resized to $32 \times 32$ pixels and treated as RGB images, to be comparable. Notice that we do not use any sample from other dataset than MNIST during training.

\begin{table}[!t]
\begin{center}
{\footnotesize
\setlength{\tabcolsep}{5pt}
\begin{tabular}{@{}rcccc@{}}
 \multicolumn{5}{c}{\textbf{Domain generalization performance of MNIST models}} \\
\toprule
& \multicolumn{4}{c}{\textbf{Testing dataset}} \\
\cmidrule(r){2-5}
\begin{tabular}{@{}c@{}}\textbf{Training} \\ \textbf{Method}\end{tabular} & \textbf{SVHN} & \textbf{SYN} & \textbf{MNIST-M} & \textbf{USPS} \\
\midrule
ERM & $.365 \pm .021$ & $.477 \pm .015$ & $.590 \pm .012$ & $.812 \pm .013$ \\

\midrule

ADA~\cite{volpi2018nips} & $.391 \pm .017$ & $.482 \pm .019$ & $.595 \pm .013$ & $.819 \pm .016$ \\

\midrule

RDA & $.395 \pm .017$ & $.603 \pm .005$ & $.751 \pm .013$ & $.832 \pm .013$ \\

\midrule

RSDA & $.474 \pm .048$ & $.620 \pm .012$ & $.815 \pm .016$ & $.831 \pm .012$ \\

\midrule

ESDA & $.489 \pm .052$ & $.622 \pm .013$ & $.816 \pm .016$ & $.840 \pm .012$ \\

\bottomrule

\end{tabular}
} 
\end{center}
\caption{\footnotesize Performance of MNIST models trained with different methods \textit{(rows)} and evaluated on test samples from different digit datasets \textit{(columns)}. Results computed by averaging over $6$ different runs.}
\label{tab:mnist_dom_gen}
\end{table}


\paragraph{Results.}

In Section~\ref{sec:test_mnist} we showed that our setup allows to find transformation tuples that lower the accuracy of MNIST models to values as low as $\sim 12\%$ (Table~\ref{tab:worst_case_mnist} -- ES row, ERM column). We are now interested in evaluating the performance on MNIST models trained through the methods detailed in Section~\ref{sec:robust_train} (RDA, RSDA and ESDA). Table~\ref{tab:worst_case_mnist} (last three columns) shows the performance of models trained with the proposed methods. The most robust model is the one trained through ESDA, for which the accuracy related to each testing case is greater than $97\%$. All our models are resistant to the negative operation applied to the images~\cite{hosseini2017negative}, with accuracy values greater than $99\%$. An important result is that there is not a statistically significant accuracy loss on original samples (Table~\ref{tab:worst_case_mnist} -- first row).

Having confirmed that we can train more robust models against the types of perturbations introduced in this work, we are interested in evaluating the performance in the domain generalization testbed; Table~\ref{tab:mnist_dom_gen} reports our findings. Also in this setting, we observe that models trained via ESDA are the most robust against distributional shifts. Models trained via RSDA are slightly less performing, but significantly more robust than the ones trained via RDA in different test cases. The more significant result is that, when testing on SVHN, there is $\sim 10\%$ gap when comparing RDA and ESDA. Furthermore, despite the transformation set $\mathbb{T}$ used is biased towards color transformations, we can observe improved performance with respect to ERM also when testing on USPS, whose samples differ from MNIST ones only in their shape. 

\subsection{CIFAR-10 Classification}

\paragraph{Experimental setup.}

We use the CIFAR-10~\cite{CIFAR} dataset, and train Wide Residual Network models (WRN~\cite{Zagoruyko2016WRN}) on the provided training set. We have chosen this class of models because they are broadly used in the community and they provide results competitive with the state of the art on CIFAR-10. We train networks with $16-$layers and set the width to $4$, choosing a trade-off between accuracy and training/testing speed, among the recipes proposed in the original work~\cite{Zagoruyko2016WRN}. We use the original code provided by the authors~\cite{WRN_code}. 

When training ERM and RDA the models, we follow the procedure proposed in~\cite{Zagoruyko2016WRN}, and run stochastic gradient descent with momentum $0.9$ for $200$ epochs, starting with a learning rate $0.1$ and decaying it at epochs $60$, $120$ and $160$. When training RSDA and ESDA models, we observed that the learning procedure is eased if the new augmentation rules are included while we are training the model with a large learning rate. For this reason, we start the learning rate decay after having searched for a satisfactory number of transformations. In the results proposed in this section, we search for $100$ different ones, and each search procedure is followed by $5$ epochs of training. 

We set the size of the tuples in $\mathbb{T}_N$ as $N=5$; with respect to the transformation set $\mathbb{T}$ described in Section~\ref{sec:search_space}, we do not include \textit{solarize} and \textit{grayscale}. When we search for transformations, we use $2,000$ samples drawn from the training set. When we test the models, we search for transformations through RS and ES using the whole test set. We run RS with $K=1,000$ iterations and three runs of ES with $K=100$ iterations; the results reported in the next paragraph are associated with the optimal $f_{min}$ found. In addition to testing the model vulnerability against such transformations, we also evaluate the domain generalization capabilities of WRN models, assessing the performance on CIFAR-10.1~\cite{recht2018cifar} dataset and on STL~\cite{pmlr-v15-coates11a} dataset. We remove samples associated with the class ``monkey'', not present in the CIFAR-10 dataset, and resize images to $32 \times 32$, to be comparable.

\paragraph{Results.}

Table~\ref{tab:worst_case_cifar10} reports the achieved results. The ``ERM'' column, which shows the results obtained by testing baseline models in different conditions, confirms the results we observed in the MNIST experiment, although with less dramatic effects. Indeed, we can find transformation tuples that the model is significantly vulnerable to, by using RS ($\sim 72\%$) and ES ($\sim 56 \%$, with a larger standard deviation). Concerning models trained with our methods, also in this experiment RDA represents an effective strategy, but RSDA and ESDA allow to train more robust models, with respect to the transformations we are testing against. 

Furthermore, the last row, reporting results obtained when testing on STL dataset, confirms the domain generalization capabilities of models trained with our method; using Algorithm~\ref{alg:robust_train}, we can observe $\sim 7\%$ improvement in accuracy, when compared against ERM. When testing on CIFAR 10.1, the benefits are less marked, but still noticeable in the RSDA case. Each accuracy value reported was obtained by averaging over $3$ different runs.

\begin{table}[!t]
\begin{center}
{\footnotesize
\setlength{\tabcolsep}{5pt}
\begin{tabular}{@{}rcccc@{}}
 \multicolumn{5}{c}{\textbf{Performance of CIFAR-10 models}} \\
\toprule
 & \multicolumn{4}{c}{\textbf{Training procedure}} \\
\cmidrule(r){2-5}
\textbf{Test} & \textbf{ERM} & \textbf{RDA} & \textbf{RSDA} & \textbf{ESDA}  \\
\midrule
Original & $.946 \pm .000$ & $.944 \pm .002$ & $.950 \pm .002$ & $.946 \pm .000$ \\
\midrule
RS & $.724 \pm .026$ & $.899 \pm .004$ & $.904 \pm .016$ & $.915 \pm .000$  \\
\midrule
ES & $.565 \pm .149$ & $.862 \pm .012$ & $.867 \pm .050$ & $.913 \pm .004$  \\
\midrule
10.1~\cite{recht2018cifar} & $.872 \pm .004$ & $.873 \pm .007$ & $.886 \pm .009$ & $.878 \pm .003$  \\
\midrule
STL~\cite{pmlr-v15-coates11a} & $.466 \pm .009$ & $.503 \pm .009$ & $.526 \pm .007$ & $.534 \pm .009$ \\

\bottomrule

\end{tabular}
} 
\end{center}
\caption{\footnotesize Performance of CIFAR-10 models trained with different methods \textit{(columns)} and tested in different conditions \textit{(rows)}. The \textit{10.1} row reports results obtained by testing on CIFAR-10.1~\cite{recht2018cifar}; the \textit{STL} row reports the ones related to STL~\cite{pmlr-v15-coates11a}. Results computed by averaging over $3$ runs.}
\label{tab:worst_case_cifar10}
\end{table}

\subsection{Semantic Scene Segmentation}

\paragraph{Experimental setup.}

We train FC-DenseNet~\cite{jegou2016tiramisu} models on the CamVid~\cite{brostow2008camvid} dataset. We use the $103-$layer version of the model, relying on an open-source implementation~\cite{semantic-segmentation-suite}. Also in this case, the choice of the model is due to its success with respect to the analyzed benchmark. The CamVid dataset contains $367$ training images, $101$ validation images and $233$ testing images from $32$ classes. We lower the sample resolution from $960 \times 720$ to $480 \times 360$, and train the models for $300$ epochs. 

When we train using RSDA/ESDA, we run RS and ES on $30$ samples from the training set, and search for new transformations every $10$ epochs. We set the size of the $N-$tuples as $N=5$. With respect to the transformation set $\mathbb{T}$ introduced in Section~\ref{sec:search_space}, we do not include \textit{solarize} and \textit{grayscale}. When we test the vulnerability of the models, we run RS (with $K=500$) and three different runs of ES (with $K=50$) on $30$ samples from the test set. As for previous experiments, we report results related to the minimum $f_{min}$ values found. Notice that the output of semantic segmentation models is richer than the output of classification models, since a prediction is associated with each pixel; indeed, $30$ samples lead to $30 \cdot 480 \cdot 360$ pixel predictions. We use pixel accuracy as a metric~\cite{FCN}. We use RMSprop~\cite{Tieleman2012} as optimizer, with decay $0.001$.

\paragraph{Results.}

Table~\ref{tab:worst_case_camvid} reports the results we obtained. They confirm the higher level of robustness of models trained via RDA, RSDA and ESDA. In this experiment though, we can observe a narrower gap between RDA and RSDA/ESDA. 

\begin{figure}[!t]
\centering
\includegraphics[width=\linewidth]{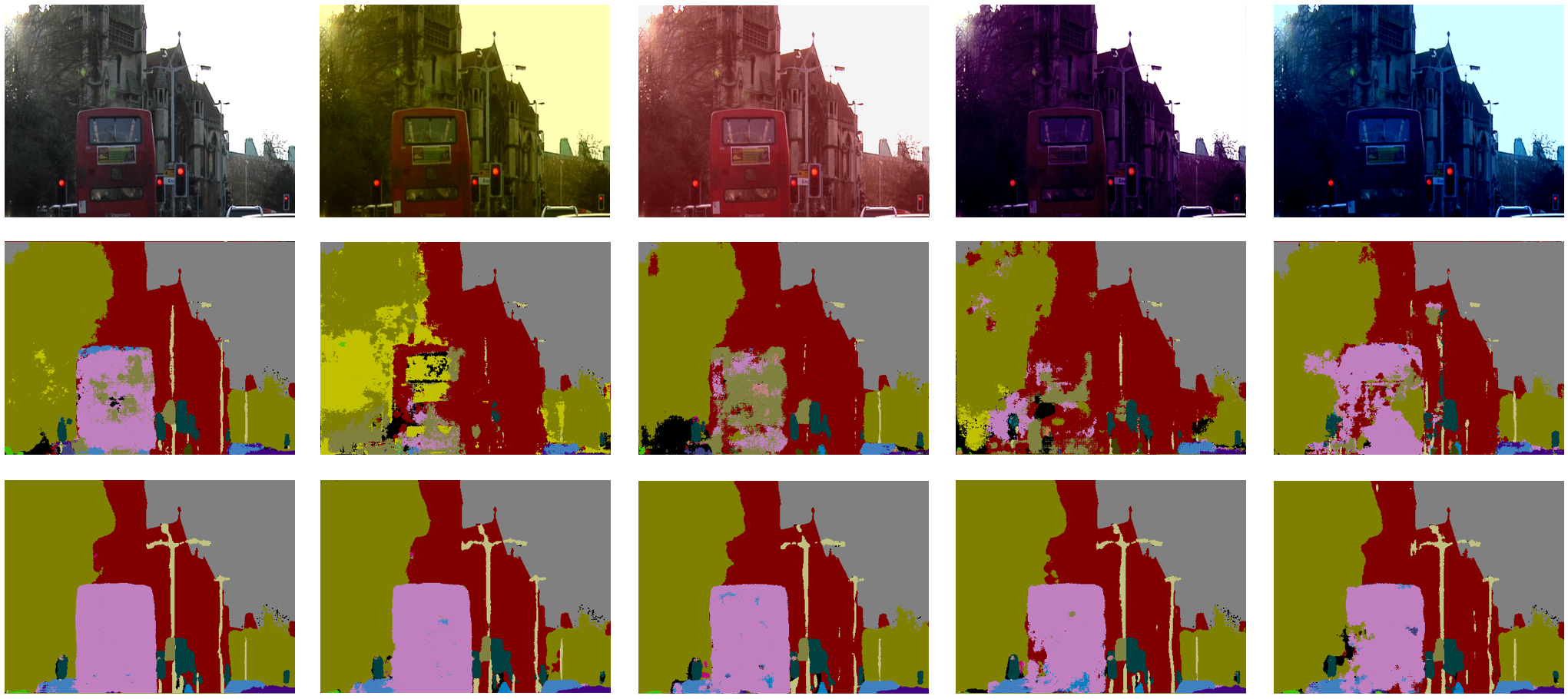}
\caption{\footnotesize A sample from CamVid \textit{(first column, first row)} modified with various image transformations. \textit{Second} and \textit{last rows} report the outputs of models trained via standard ERM and via Algorithm~\ref{alg:robust_train}, respectively.}
\label{fig:camvid}

\end{figure}

Figure~\ref{fig:camvid} shows the output of a model trained via ESDA (middle) and the output of a model trained via standard ERM (bottom), when the original input (first column, top) is perturbed with different image transformations (top). These results not only qualitatively show the better performance of ESDA, but also that the transformation tuples we are sampling from $\mathbb{T}_N$ are realistic approximations of possible visual conditions that a vision module (for instance, for a self-driving car) might encounter. For example, images in the middle row, second and third column, can be considered as simulations of the light conditions that one could encounter during dawn or sunset---and in which the baseline model performs poorly. 

\begin{table}[!t]
\begin{center}
{\footnotesize
\setlength{\tabcolsep}{5pt}
\begin{tabular}{@{}rcccc@{}}
 \multicolumn{5}{c}{\textbf{Performance of CamVid models}} \\
\toprule
& \multicolumn{4}{c}{\textbf{Training procedure}} \\
\cmidrule(r){2-5}
\textbf{Test} & \textbf{ERM} & \textbf{RDA} & \textbf{RSDA} & \textbf{ESDA}  \\

\midrule
Original & $.862 \pm .007$ & $.851 \pm 004$ & $.854 \pm 003$ & $.851 \pm .002$ \\

\midrule
RS & $.458 \pm .027$ & $.812 \pm .007$ & $.825 \pm .009$ & $.820 \pm .007$ \\

\midrule
ES & $.311 \pm .013$ & $.811 \pm .011$ & $.824 \pm .008$ & $.822 \pm .008$ \\

\bottomrule

\end{tabular}
} 
\end{center}
\caption{\footnotesize Performance of CamVid models trained with different methods \textit{(columns)} and tested in different conditions \textit{(rows)}. Results computed by averaging over $3$ different runs.}
\label{tab:worst_case_camvid}
\end{table}

\subsection{Face detection}

\paragraph{Experimental setup.}

We test our search methods on a widely used API for face detection~\cite{face-recognition-api}, that takes RGB images as input and provides in output the locations of the faces in the image. We use four subsets of $1,000$ images uniformly sampled from the MS-Celeb-1M~\cite{guo2016msceleb}, resized to $64 \times 64$, as datapoints in input to RS and ES. Each image contains one celebrity face, thus the API gives in output a single location if it detects a face, or nothing otherwise. In practical terms, due to the nature of the input, we can interpret the API as a binary model and test it through the optimization problem~\ref{eq:worstcase_class}. We set the number of transformations as $N=3$ and $N=5$. For each $N$ value and each subset of faces, we run RS with $K=15,000$ iterations, and run $6$ different runs of ES with $K=100$. We average results over the optimal $f_{min}$ values obtained in the four subsets. With respect to the transformation set $\mathbb{T}$ depicted in Section~\ref{sec:search_space}, we do not include \textit{solarize}.

\paragraph{Results.}

\begin{figure}[t]
\centering
\includegraphics[width=\linewidth]{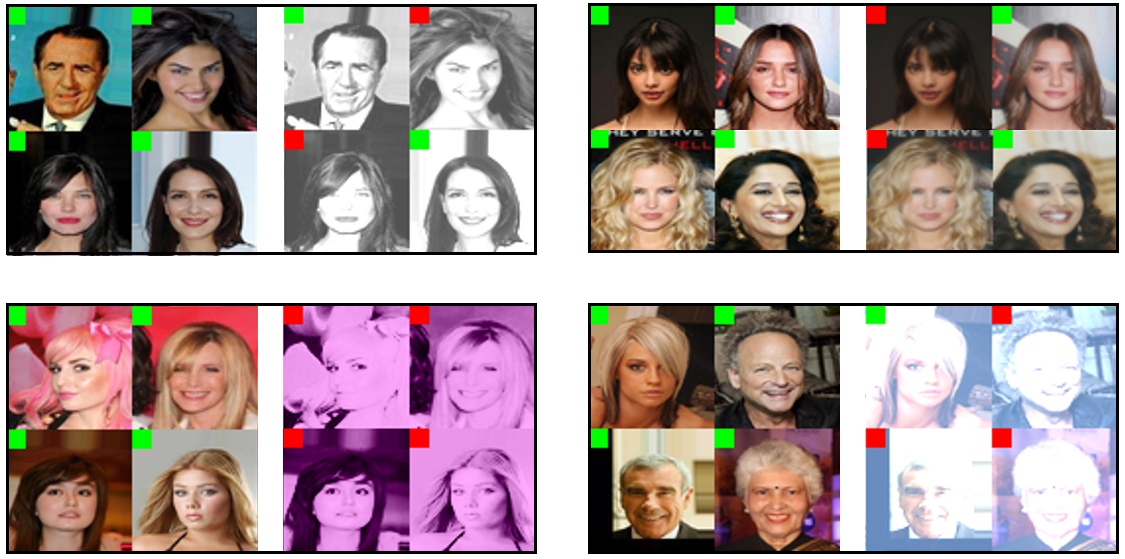}
\caption{\footnotesize Four different examples of transformations found via RS and ES. Green and red squares indicate whether the API has detected the face or not, respectively.}
\label{fig:res_faces}

\end{figure}

Table~\ref{tab:worst_case_face} reports the accuracy values obtained, and Figure~\ref{fig:res_faces} reports different examples of faces modified through the transformation tuples found via RS and ES. Green and red squares indicate whether the API has detected or not a face, respectively. Qualitatively, we observed that the model tends to fail when the input manipulation is such that some facial features are no longer visible or deteriorated (for example, the nose). The importance of these vulnerabilities depends on the different API use cases. For example, vulnerability to some grayscale tones might not matter for a model that deals with images recorded in the streets, but it might matter for a social network application. Vulnerability to extreme brightness conditions can be harmful for a street camera, where the broad variety of possible visual conditions might not allow to have a proper view of the facial features. One strength of the search methods we proposed is that they allow users to set transformation sets according to the applications they are concerned about.

\begin{table}[!t]
\begin{center}
{\footnotesize
\setlength{\tabcolsep}{2.5pt}
\begin{tabular}{@{}rcccc@{}}
 \multicolumn{5}{c}{\textbf{Performance of Face Detection API~\cite{face-recognition-api}}} \\
\toprule

\begin{tabular}{@{}c@{}}\textbf{Original} \\ \textbf{Accuracy}\end{tabular} &  
\begin{tabular}{@{}c@{}}\textbf{RS} \\ \textbf{N=3}\end{tabular} &  
\begin{tabular}{@{}c@{}}\textbf{ES} \\ \textbf{N=3}\end{tabular} &  
\begin{tabular}{@{}c@{}}\textbf{RS} \\ \textbf{N=5}\end{tabular} &  
\begin{tabular}{@{}c@{}}\textbf{ES} \\ \textbf{N=5}\end{tabular}   \\  
\midrule

$.878 \pm .007$ & $.789 \pm .010$ & $.705 \pm .069$ & $.516 \pm .014$ &  $.174 \pm .134$ \\

\bottomrule

\end{tabular}
} 
\end{center}
\caption{\footnotesize Accuracy of the Face Detection API in different testing conditions, analogously to the previous tests reported in this work.}
\label{tab:worst_case_face}
\end{table}

\section{Conclusions}\label{sec:conclusions}

We propose a combinatorial optimization problem to find distributional shifts that a given model is vulnerable to, in terms of $N-$tuples of image transformations. We show that random search and, in particular, evolution-based search are effective approaches to face this problem. 
Further, we show that the same search algorithms can be exploited in a training procedure, where harmful distributional shifts are searched and harnessed. 
We report results for a variety of tasks (classification, segmentation and face detection), showing that the problem formulation is flexible and can be adopted in different circumstances.

Among others, some valuable directions for future works consist in (i) the implementation of more effective methods to approach the optimization problem~\ref{eq:worstcase}, in order to find more harmful transformations with reduced computational cost, (ii) the analysis of more complex transformation sets, and (iii) the definition of a proper content-preserving constraint in the optimization problem. 

\paragraph*{Acknowledgments.}

We are grateful to Jacopo Cavazza and Federico Marmoreo for helpful discussions concerning the problem formulation proposed in this work. 

\clearpage

{\small 
\bibliographystyle{ieee}
\bibliography{egbib}
}

\clearpage

\appendix

\onecolumn

\section{Transformation Sets}\label{appendixA}

\begin{table}[ht]
\begin{center}
{\normalsize
\setlength{\tabcolsep}{5pt}
\begin{tabular}{@{}rccccccc@{}}
 \multicolumn{7}{c}{\textbf{Transformation Sets $\mathbb{T}$ for the Experiments}} \\
\toprule
& & & \multicolumn{4}{c}{\textbf{Experiment}} \\
\cmidrule(r){4-7}
\textbf{Transformations} & \textbf{Range} & \textbf{No. Levels} & \textbf{MNIST~\cite{MNIST}} & \textbf{CIFAR-10~\cite{CIFAR}} & \textbf{CamVid~\cite{brostow2008camvid}} & \textbf{Faces~\cite{face-recognition-api}} \\
\midrule
\textit{Autocontrast}                & $[0.0,0.3]$ & $20$ & \checkmark & \checkmark & \checkmark & \checkmark \\
\midrule
\multirow{2}{*}{\textit{Brightness}} & $[0.6,1.4]$ & $20$ & \checkmark & & &  \\
                                     & $[0.8,1.2]$ & $20$ & & \checkmark & \checkmark & \checkmark\\
\midrule
\textit{Color}     & $[0.6,1.4]$     & $20$ & \checkmark & \checkmark & \checkmark & \checkmark\\
\midrule
\textit{Contrast}  & $[0.6,1.4]$     & $20$ & \checkmark & \checkmark & \checkmark & \checkmark\\
\midrule
\textit{Sharpness} & $[0.6,1.4]$     & $20$ & \checkmark & \checkmark & \checkmark & \checkmark\\
\midrule
\textit{Solarize}  & $[0.0,20.0]$    & $20$ & \checkmark & & & \\
\midrule
\textit{Grayscale} & $-$      & $1$  & \checkmark & & & \checkmark\\
\midrule
\multirow{2}{*}{\textit{R-channel enhancer}}& $[-120,120]$ & $30$ & \checkmark & & \checkmark & \checkmark\\
                                            & $[-30,30]$ & $30$ & & \checkmark &  & \\
\midrule
\multirow{2}{*}{\textit{G-channel enhancer}}& $[-120,120]$ & $30$ & \checkmark & & \checkmark & \checkmark\\
                                            & $[-30,30]$ & $30$ &  & \checkmark &  & \\
\midrule
\multirow{2}{*}{\textit{B-channel enhancer}}& $[-120,120]$ & $30$ & \checkmark & & \checkmark & \checkmark\\
                                            & $[-30,30]$ & $30$ &  & \checkmark & & \\
\midrule
\midrule

\textbf{Size of $\mathbb{T}_N$} & & & $211^N$ &  $190^N$ &  $190^N$ &  $191^N$ \\  

\bottomrule

\end{tabular}
} 
\end{center}
\caption{\footnotesize List of transformations used.}
\label{tab:set}
\end{table}


In this Section, we report the image transformations briefly introduced in Section~2.1, and used throughout Sections~3.3 and 5. Table~\ref{tab:set} reports them (column $1$), with the range of magnitude values (column $2$) and the number of values in which the ranges have been discretized (column $3$). Columns $4-7$ indicate whether a transformation is used (\checkmark) in the experiments of Sections~5.1, 5.2, 5.3 and 5.4, respectively. Grayscale conversion (\textit{``Grayscale''} row) has only one magnitude value. For all the transformations, excluding the \textit{R/G/B-channel enhancement} operations, the reader can refer to the PIL library~\cite{pil-library}, and in particular to the modules~\cite{pillow-image-ops,pillow-image-enhance}. We summarize them in the following, reporting the core descriptions.

\begin{itemize}

    \item \textbf{Autocontrast:} ``Maximize (normalize) image contrast. This function calculates a histogram of the input image, removes cutoff percent of the lightest and darkest pixels from the histogram, and remaps the image so that the darkest pixel becomes black (0), and the lightest becomes white (255)''~\cite{pillow-image-ops}.
    
    \item \textbf{Brightness:} ``Adjust image brightness. [...] An enhancement factor of 0.0 gives a black image. A factor of 1.0 gives the original image''~\cite{pillow-image-enhance}.
    
    \item \textbf{Color:} ``Adjust image color balance. [...] An enhancement factor of 0.0 gives a black and white image. A factor of 1.0 gives the original image''~\cite{pillow-image-enhance}.
    
    \item \textbf{Contrast:} ``Adjust image contrast. [...] An enhancement factor of 0.0 gives a solid grey image. A factor of 1.0 gives the original image''~\cite{pillow-image-enhance}.
    
    \item \textbf{Sharpness:} ``    Adjust image sharpness. [...] An enhancement factor of 0.0 gives a blurred image, a factor of 1.0 gives the original image, and a factor of 2.0 gives a sharpened image''~\cite{pillow-image-enhance}.
    
    \item \textbf{Solarize:} ``Invert all pixel values above a threshold''~\cite{pillow-image-ops}.
    
    \item \textbf{Grayscale:} ``Convert the image to grayscale''~\cite{pillow-image-ops}. We treat the output as an RGB image by replicating it in three different channels. 
    
    \item \textbf{R-channel enhancer:} add a value to the R-channel of all pixels.
    
    \item \textbf{G-channel enhancer:} add a value to the G-channel of all pixels
    
    \item \textbf{B-channel enhancer:} add a value to the B-channel of all pixels

\end{itemize}

\end{document}